\documentclass[letterpaper, 10pt, conference]{ieeeconf}  

\IEEEoverridecommandlockouts                              
\usepackage[caption=false,font=footnotesize]{subfig}
\usepackage{graphics} 
\usepackage{epsfig} 
\usepackage{mathptmx} 
\usepackage{times} 
\usepackage{amsmath} 
\usepackage{amssymb}  
\usepackage{multicol}
\usepackage{tabularx}
\usepackage[normalem]{ulem}
\usepackage{multirow}
\usepackage{tikz}
\usepackage{booktabs}
\usepackage{balance}
\usepackage[letterpaper, left=0.75in, right=0.75in, bottom=0.75in, top=0.75in]{geometry}

\title{\LARGE \bf
Automated Lane Change Strategy using Proximal Policy Optimization-based Deep Reinforcement Learning 
}

\author{Fei Ye$^{*1}$, Xuxin Cheng$^{*1,2}$, Pin Wang$^{1}$, Ching-Yao Chan$^{1}$ and Jiucai Zhang$^{3}$
\thanks{$^{*}$ Equal contribution.}
\thanks{$^{1}$ F. Ye, X. Cheng, P. Wang and C. Chan are with California PATH, University of
California, Berkeley, Richmond, CA, 94804, USA. 
        {\tt\small \{fye,  chengxuxindavid, pin\_wang, cychan\}@berkeley.edu.}}%
\thanks{$^{2}$ X. Cheng is also with Beijing Institute of Technology, Beijing, 100081, China.}%

\thanks{$^{3}$ J. Zhang is with GAC R\&D Center Silicon Valley Inc., 639 N. Pastoria Ave, Sunnyvale, CA, 94085, USA}
}

\begin{document}

\maketitle
\thispagestyle{empty}
\pagestyle{empty}

\begin{abstract}
Lane-change maneuvers are commonly executed by drivers to follow a certain routing plan, overtake a slower vehicle, adapt to a merging lane ahead, etc. However, improper lane change behaviors can be a major cause of traffic flow disruptions and even crashes. While many rule-based methods have been proposed to solve lane change problems for autonomous driving, they tend to exhibit limited performance due to the uncertainty and complexity of the driving environment. Machine learning-based methods offer an alternative approach, as Deep reinforcement learning (DRL) has shown promising success in many application domains including robotic manipulation, navigation, and playing video games. However, applying DRL to autonomous driving still faces many practical challenges in terms of slow learning rates, sample inefficiency, and safety concerns. In this study, we propose an automated lane change strategy using proximal policy optimization-based deep reinforcement learning, which shows great advantages in learning efficiency while still maintaining stable performance. The trained agent is able to learn a smooth, safe, and efficient driving policy to make lane-change decisions (i.e. when and how) in challenging situation such as dense traffic scenarios. The effectiveness of the proposed policy is validated by using metrics of task success rate and collision rate. The simulation results demonstrate the lane change maneuvers can be efficiently learned and executed in a safe, smooth and efficient manner.
\end{abstract}

\section{Introduction}
Automated and semi-automated vehicles are considered to have a great potential to improve transportation safety and efficiency, and a considerable amount of studies has been performed with a focus on autonomous driving or advanced driving assistance systems (ADAS). However, even with progress to date, it is still challenging to achieve automated decision-making and control to drive vehicles effectively and safely in some cases. These cases often occur in a dynamically changing environment, which involves complex interactions with the surrounding objects. While expert demonstrations from human drivers can be used to train algorithms to learn certain driving tasks via imitation learning \cite{NVIDIA}, a substantial amount of data needs to be collected at all possible conditions (with variation in surrounding traffic, road signs, traffic light) for training, which is costly and often impractical. Moreover, it requires massive human labor to label such data, but still may not cover all of the complex situations in real-world driving scenarios. 

On the other hand, reinforcement learning (RL) algorithms have shown great potential for handling decision-making and control problems. It can learn the task in a trial-and-error way that does not require explicit human labeling or supervision on each data sample. RL has demonstrated significant success for solving complex task in both robotic manipulations \cite{finn2016guided} and playing video games \cite{Atari2013}.
%
%

However, applying RL to real-world applications is particularly challenging, especially for autonomous driving tasks that involve extensive interactions with other vehicles in a dynamically changing environment. Among a variety of vehicle decision-making and control problems that were tackled using RL algorithms \cite{wang2019continuous, review-17, wang2019quadratic}, facilitating automated lane change maneuvers is of special interests, since improper lane change behaviors can be a major cause of highway crashes and traffic jams \cite{Collision1, Collision2, LHP, CooperativeLC-19}. An early work of applying deep RL to lane change can be found in \cite{QDRL-17}, where the Q-masking technique is proposed to act as a mask on the output Q-values in a deep Q-learning framework to obtain a high-level policy for tactical lane change decisions, while still maintaining a tight integration with the prior knowledge, constraints and information from a low-level controller. To overcome the deficiencies of rule-based models that are prone to failures in unexpected situations or diverse scenarios, a Q-function approximator with closed form greedy policy is proposed in \cite{Pin-2018}. Further improvement of vehicle continuous control has been made in \cite{wang2019quadratic} to tackle more challenging situation. In \cite{Tianyu-19}, a hierarchical RL based architecture is presented to make lane change decisions and execute control strategies. Specifically, a deep Q-network (DQN) is trained to decide when to conduct the maneuver based on safety considerations, while a deep Q-learning framework with a quadratic approximator is designed to complete the maneuver in the longitudinal direction. 

Additionally, the applications of deep RL algorithms on lane change strategies are often burdened by their slow learning rates. In \cite{HighLC-18}, this problem is addressed by making use of a minimal state representation consisting of only 13 continuous features, which facilitates a faster learning rate while training a DQN. Moreover, a technique referred to as ``multi-objective approximate policy iteration (MO-API)'' is presented in \cite{AutoDM-19}. The value and policy approximations are learned using data-driven feature representations, where sparse kernel-based features or manifold-based features can be constructed based on data samples. It is concluded that higher learning efficiency can be achieved using the proposed MO-API approach, when compared to benchmark RL algorithms such as multi-objective Q-learning. While these methods are centered around manipulating the feature space, more effective RL algorithms can also be employed to facilitate fast learning rates and reduce the high variance in policy learning. Moreover, the safety aspect of autonomous driving is also critical. Although deep reinforcement learning methods can maximize the designed reward, it does not necessarily guarantee safety during learning or execution phases \cite{Shield2017}. 

To address these aforementioned issues, in this work, we thus propose a safe proximal policy optimization (PPO)-based deep reinforcement learning method, which combines the policy with a safety intervention module. Built upon an actor-critic structure, the parameterized actor of PPO \cite{PPO} can enforce a trust region with clipped objectives, which, when compared to the classical TRPO \cite{TRPO}, can reduce the burden in computing nonlinear conjugate gradients. Moreover, PPO tends to be more efficient in sampling and learning policies than TRPO. Meanwhile, when compared to value-based methods, PPO is able to compute actions directly from the policy gradient, rather than from optimizing the value function. On the other hand, the merit of the critic is to supply the actor with the knowledge of performance in low variance. All of these nice properties of PPO can improve its capability in real-life applications. To further enhance safety in both learning and execution phases, a safety intervention module \cite{SafeRL2017} is added to reduce the chance of taking catastrophic actions in a complex interactive environment. Moreover, we introduce a novel lateral action for decision making -- aborting lane change, which enables our ego-vehicle to avoid potential collisions by \emph{aborting} and changing back to the original lane at any point while undertaking the lane change action. In the longitudinal direction, the ego vehicle needs to choose which leading vehicle to follow, so it can perform speed adjustments even before making the actual lane change.
%

%
%

The objective of this study is to develop an decision-making strategy to enable automated mandatory lane change maneuvers aiming at achieving the objectives of \emph{safety}, \emph{efficiency}, and \emph{comfort}, using PPO-based deep reinforcement learning. The technical preliminaries of the PPO algorithm are introduced in Section II. Section III presents a detailed problem description, while the formulation of the proposed lane change model and reward function design are described in Section IV. The simulation setup, evaluation metrics, and results are presented in Section V. Section VI gives the conclusion and future work directions.
\begin{figure*}[!t]
\centering
\includegraphics[width=5.6in]{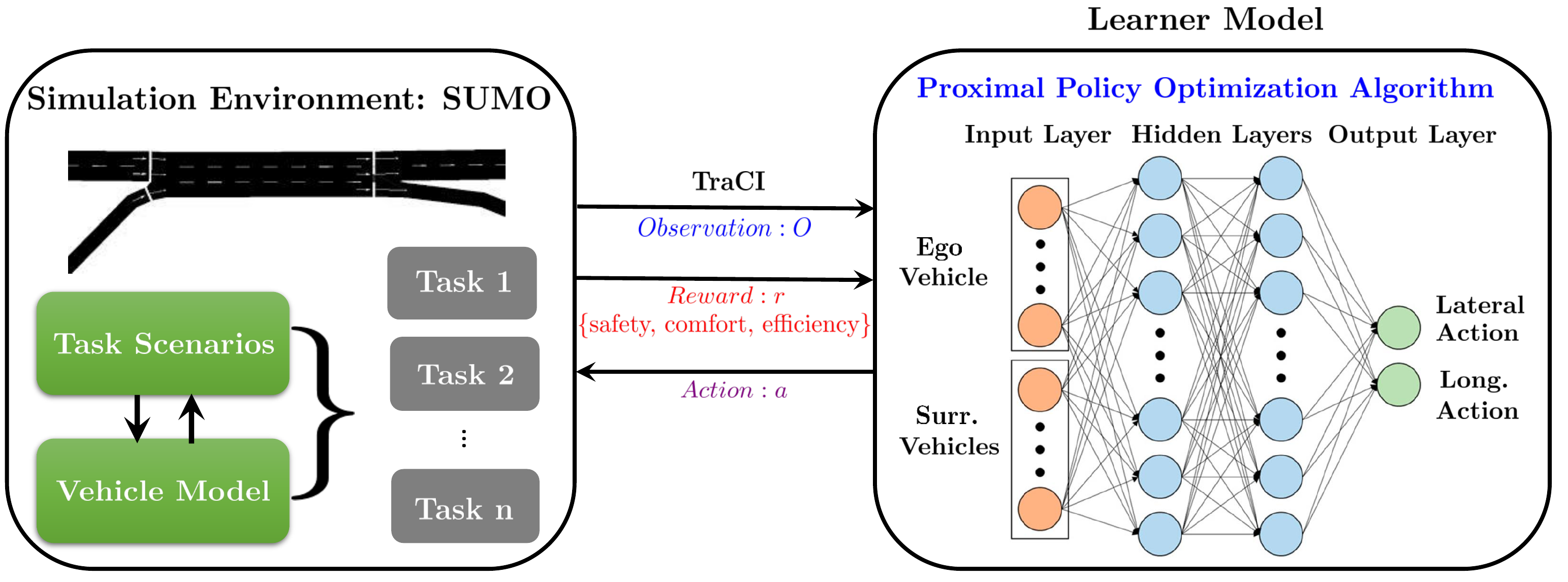}
\caption{System architecture of the proposed PPO-based lane change method}
\label{fig:framework}
\end{figure*}
\section{Policy-based Reinforcement Learning}
Reinforcement learning can teach how an \emph{agent} to act by interacting with its \emph{environment} in order to maximize the expected cumulative rewards for a certain task. There are two general categories of RL algorithms, namely value-based method and policy-based method. While value-based methods can approximate the value function using neural networks in an off-policy way, the primary advantage of policy-based methods, such as the REINFORCE algorithm \cite{Reinforce}, is that they can directly optimize the quantity of advantage, while remaining stable during function approximations. Our study thus focuses on policy-based RL methods. 

The loss function for updating a RL policy has the general form as
\begin{equation}
L^{PG}(\theta)=\hat{\mathbb{E}}_{t}\left[\log \pi_{\theta}\left(a_{t} | s_{t}\right) \hat{A}_{t}\right]
\end{equation}
where $\mathbb{E}_{t}$ is the expectation operator, $\pi_{\theta}$ is a stochastic RL policy, $\hat{A}_{t}$ is an estimator of the advantage function at time step $t$.

While it might be appealing and straightforward to perform multiple steps of optimization on this loss function $L^{PG}(\theta)$, many challenges can arise from the prevalence of sample inefficiency, the balance between exploration and exploitation, and the undesirable high variance of the learned policy. Empirically, it often leads to destructively large policy updates, which are destructive since they can also affect the observation and reward distribution at future time steps. 

Compared to this fundamental loss function $L^{PG}(\theta)$ to update a policy, some advanced policy-based algorithms such as TRPO \cite{TRPO} and PPO \cite{PPO} take the actor-critic structure, which is able to combine advantages of traditional value-based and policy-based approaches. The PPO algorithm is also much simpler to implement, has less computation burden, and has better sample complexity (empirically). Specifically, PPO proposes a clipped surrogate loss function and combines the policy surrogate and a value function error term, which is defined as \cite{PPO}
\begin{equation}
L_{t}^{CLIP+VF+S}(\theta)=\hat{\mathbb{E}}_{t}\left[L_{t}^{CLIP}(\theta)-c_{1} L_{t}^{V F}(\theta)+c_{2} S\left[\pi_{\theta}\right]\left(s_{t}\right)\right]
\end{equation}
where $L_{t}^{CLIP}$ is the clipped surrogate objective, $c_1$, $c_2$ are coefficients, $L_{t}^{VF}$ is the squared-error loss of the value function $(V_{\theta}\left(s_{t}\right)-V_{t}^{\operatorname{targ}})^{2}$, and $S$ denotes the entropy loss. Specifically, the clipped surrogate objective $L_{t}^{CLIP}$ takes the following form
\begin{equation}
L^{CLIP}(\theta)=\hat{\mathbb{E}}_{t}\left[\min \left(r_{t}(\theta) \hat{A}_{t}, \operatorname{clip}\left(r_{t}(\theta), 1-\epsilon, 1+\epsilon\right) \hat{A}_{t}\right)\right]
\end{equation}
where $\epsilon$ is a hyperparameter, and $r_{t}(\theta)$ denotes the probability ratio $r_{t}(\theta)=\pi_{\theta}\left(a_{t}|s_{t}\right)/\pi_{\theta_{\text{old}}}\left(a_{t}| s_{t}\right)$. In this manner, the probability ratio $r$ is clipped at $1-\epsilon$ or $1 +\epsilon$ depending on whether the advantage is positive or negative, which forms the clipped objective after multiplying the advantage approximator $\hat{A}_{t}$. The final value of $L_{t}^{CLIP}$ takes the minimum of this clipped objective and the unclipped objective $r_{t}(\theta) \hat{A}_{t}$, which can effectively avoid taking a large policy update compared to the unclipped version \cite{PPO}, which is also known as the loss function of the conservative policy iteration algorithm \cite{CPI}.

A complete PPO algorithm uses a fixed length-$T$ trajectory segments at each iteration, which runs the policy for $T$ times steps, and each of the $N$ parallel actors will collect data in all time steps. A truncated version of generalized advantage estimation is adopted, which takes the form of
\begin{equation}
\hat{A}_{t}=\delta_{t}+(\gamma \lambda) \delta_{t+1}+\cdots+\cdots+(\gamma \lambda)^{T-t+1} \delta_{T-1}
\end{equation}
where $\delta_{t}=r_{t}+\gamma V\left(s_{t+1}\right)-V\left(s_{t}\right)$, and $\gamma$ is the discount factor. 

Then PPO constructs the loss in Eqn. (2) on these $NT$ time steps of data, and optimize it with mini-batch SGD, for $K$ epochs \cite{PPO}.

%


%

%
%

\section{Problem Description}
A lane change action is usually conducted when the vehicle needs to exit the highway, overtake a slower vehicle, adapt to the merging lane ahead, etc. Lane change maneuvers can be classified into two major categories: mandatory and discretionary. Compared to a discretionary lane change that is intended to achieve faster speed or better driving experience, a mandatory lane change is usually occurs when the ego vehicle is forced to make a lane change due to either a lane drop or a highway exit.

In our formulation, we focus on mandatory lane change situations where explicit lane change intentions are already given by the vehicle route planner, and our task is to decide when and how to make the lane change maneuver based on states of surrounding vehicles and the ego vehicle itself. Once a decision is made by the model, a low-level controller  responsible for regulating longitudinal and lateral movements is used to generate a corresponding control command to execute the decision. The applicable lane change policy to be learned should incorporate the following three functionalities:
\begin{itemize}
  \item Avoiding collisions with surrounding vehicles
  \item Achieving high driving efficiency
  \item Executing smooth maneuvers
\end{itemize}
\section{METHODOLOGY}
This study proposes a PPO-based RL approach for learning a robust and reliable mandatory lane change strategy. In this section, we'll introduce the system architecture, the design of state space, action space, and reward functions of our proposed decision-making strategy.
\subsection{System Architecture}
The overall system architecture for enabling automated lane change is shown in Fig. \ref{fig:framework}. There are two major components in the system: a learner model and a simulation environment. Specifically, the learner model uses PPO to train the ego-vehicle (agent) to learn a high-level policy for decision making tasks while interacting with the surrounding traffic. The simulation environment, which includes the road network, traffic, and different task scenarios, is generated using a high-fidelity microscopic traffic simulation suite SUMO (Simulation of Urban Mobility) \cite{SUMO}, and it is used to interact with the training agent. Using SUMO and its associated traffic control interface (TraCI), we can access the vehicle information in the road network, and execute high-level decisions in the learner model and take into account vehicle dynamics generated in the simulation model. 

To enable safe, smooth, and efficient driving behaviors on highways, the ego-vehicle first receives its current state and its surrounding vehicles' state from the SUMO environment through TraCI, and these states are passed through the policy network. Next, the ego-vehicle determines the  high-level longitudinal and lateral actions based on the developed policy network, which then sends the action back to SUMO to model the vehicle's movement in the next time step and compute the corresponding reward.
%
%
%
\begin{table}[]
    \caption{Defined Action Space}
    \resizebox{\linewidth}{!}{
\begin{tabular}{lcc}
\toprule
\multirow{2}{*}{Lateral command} & \multicolumn{2}{c}{\begin{tabular}[c]{@{}c@{}} Logitudinal command \\ (choosing which leading vehicle to follow) \end{tabular} }  \\ \cmidrule{2-3} 
& 0: Current lane leader     & 1: Target lane leader       \\ \midrule
0: Lane keeping                            & \checkmark         & \checkmark       \\
1: Changing to the target lane             & \checkmark         & \checkmark       \\
2: Aborting lane change                    & \checkmark         & \checkmark       \\
\bottomrule
\end{tabular}}
\label{tab:action_space}
\end{table}
\subsection{State and Action Space}
\begin{figure}[!t]
\centering
\includegraphics[width=3.0in]{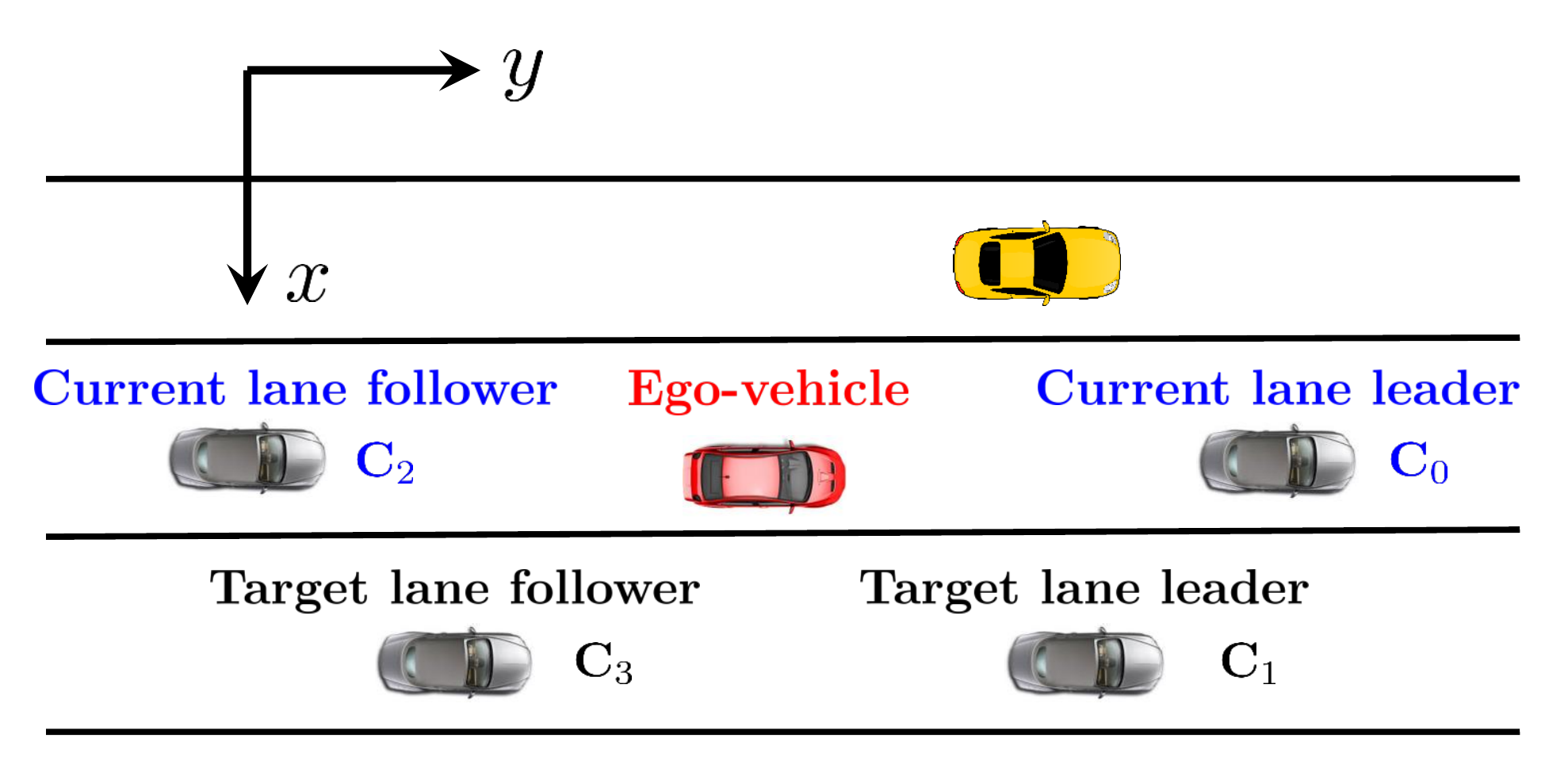}
\caption{Vehicle lane change scenario}
\label{fig:vehicles}
\end{figure}

We consider the state of 5 vehicles involved in the lane change decision and execution phase as shown in Fig. \ref{fig:vehicles}: (1) the ego-vehicle $C_e$; (2) the leading vehicle in the current lane $C_0$; (3) the leading vehicle in the target lane $C_1$; (4) the following vehicle in the current lane $C_2$; and (5) the following vehicle in the target lane $C_3$. 
\begin{table*}[]
    \caption{Near-Collision Reward Conditioned on the Action Space}
    \resizebox{\linewidth}{!}{
\begin{tabular}{lcc}
\toprule
$R_{near\_collision}$ & $C_0$: Current lane leader    $C_2$: Current lane follower  & $C_1$: Target lane leader      $C_3$: Target lane follower  \\ \midrule
Lateral action 0: Lane keeping                            & $F(C_e,C_1)$         & 0       \\
Lateral action 1: Changing to the target lane             & 0         & $\min(F(C_e,C_1), F(C_e,C_3))$       \\
Lateral action 2: Aborting lane change                    & $\min(F(C_e,C_0), F(C_e,C_2))$         & 0       \\
\bottomrule
\end{tabular}}
\label{tab:reward}
\end{table*}

The state space is composed of a total of 21 continuous state variables from both the ego-vehicle and its surrounding vehicles. Specifically, the ego-vehicle has 5 state variables, including its longitudinal position, speed, acceleration, as well as its lateral position and speed. Additionally, each surrounding vehicle has 4 state variables: relative distance to the ego-vehicle, longitudinal speed, acceleration, and lateral position.  

In this study, we design the action space in both the lateral and longitudinal direction, so that an agent can learn when and how to perform a lane change. For the lateral command, we have three different actions \{0, 1, 2\}, in which ``0'' denotes lane keeping, ``1'' represents making lane change right away, and ``2'' indicates aborting the lane change maneuver. In a real-world scenario, a driver's execution of the lane change decision can also be affected by the interactions between the ego vehicle and other vehicle. For instance, the driver can abort the lane change decision if another vehicle is merging to the same spot simultaneously. Therefore, we've added an aborting-lane-change action to the lateral action space, which enables the vehicle to abort taking a lane change action in case of a potential collision.
For the longitudinal strategy, there are two actions \{0, 1\}, in which each chosen action denotes whether to follow the current lane leader ``0'', or the target lane leader ``1'' to adjust the vehicle to a proper longitudinal position.

Therefore, there are altogether 6 actions combining different cases of actions in both the lateral and longitudinal direction, which are defined as shown in Table \ref{tab:action_space}. We also designed the low-level longitudinal and lateral controller to execute the corresponding movement instructed by the high-level PPO decision maker. To generate a human-like smooth trajectory, we modified a well-developed car-following intelligent driver model (IDM) \cite{IDM} to suit our needs in this low-level longitudinal control. 
\subsection{Reward Function}
%
The reward function is designed to incorporate key objectives of this study, which is to develop an automated lane change strategy centered around \emph{safety}, \emph{efficiency}, and \emph{comfort}. More specifically, these ideas are explained as follows:
\begin{enumerate}
    \item \emph{Comfort}: evaluation of jerk (lateral and longitudinal direction);
    \item \emph{Efficiency}: evaluation of travel time and relative distance to the target lane;
    \item \emph{Safety}: evaluation of the risk of collisions and near-collisions.
\end{enumerate}

The reward function representing comfort can be expressed as
\begin{equation}
    R_{comf}(t) = -\alpha \cdot \dot{a}_x(t)^2 - \beta \cdot \dot{a}_y(t)^2
\end{equation}
where $\dot{a}_x$ and $\dot{a}_y$ are the lateral jerk and the longitudinal jerk; and $\alpha$ and $\beta$ are the corresponding weights for lateral and longitudinal comfort. This reward function is introduced to avoid sudden acceleration or deceleration of the vehicle that may cause vehicle occupant discomfort.

In terms of efficiency, the ego-vehicle should manage to move to the target lane as soon as possible without exceeding the speed limit, thus the efficiency reward function can be defined as
\begin{equation*}
\begin{cases}
   R_{time}(t)  =  -\delta t \\
   R_{lane}(t)  =  -|P_x - P_{x}^{*}| \\
   R_{speed}(t) = -|V_y - V_{desired}|
\end{cases}
\end{equation*}
\begin{equation}
    R_{eff}(t) = w_t \cdot R_{time}(t) + w_l \cdot R_{lane}(t) + w_s \cdot R_{speed}(t)
\end{equation}
where $R_{time}(t)$ is the sub-reward related to time, $R_{lane}(t)$ represents the difference between the ego-vehicle's lateral position $P_x$ with respect to the targeted lateral position $P_{x}^{*}$, while $R_{speed}(t)$ represents the difference between the ego-vehicle's longitudinal speed $P_x$ with respect to the targeted longitudinal speed $V_{desired}$. The total efficiency reward $R_{eff}(t)$ can be computed as the sum of these three rewards adjusted by their corresponding weights $w_t$, $w_l$, and $w_s$, respectively. 

In terms of safety, we introduce a near-collision penalty term $R_{near\_collision}$ conditioned on the action-state space, rather than only penalizing a collision that actually takes place. In this case, an ego-vehicle $C_e$ can learn to abort the lane-change maneuver if its relative distances to the surrounding vehicles $C_i$ are smaller than a predefined threshold, indicating a collision is likely to happen. The specific form of this near-collision penalty term $R_{near\_collision}$ in terms of their relative positions is shown in TABLE \ref{tab:reward}, in which $F(C_e,C_i)$ is defined as
\begin{equation}
    F(C_e,C_i)=-1/(|P_{y\_e}-P_{y\_i}|+0.1)
\end{equation}
where $P_{y\_e}$ represents the longitudinal position of the ego-vehicle $C_e$; and $P_{y\_i}$ represents the longitudinal position of the surrounding vehicle $C_i$.
%

The penalty for collision and near-collision takes the form of
\begin{equation} 
R_{collision} = 
\begin{cases} R_{near\_collision}   & \mathrm{if}\ D<d_s\\
-100  &  \mathrm{if}\ \mathrm{collision} 
\label{eqrwd}
\end{cases}
\end{equation}

Additionally, in order to enhance safety and distinguish catastrophic actions and sub-optimal actions in the automated lane changing process, a safety intervention module is designed to label the output action from the actor network as ``catastrophic'' or ``safe''. The safety intervention module can then replace these ``catastrophic'' actions with safer actions and return a negative penalty reward $R_{p}(t)$.

Therefore, the complete form of safety reward is
\begin{equation} 
R_{safety} = R_{collision} + R_{p}
\label{eqsafety}
\end{equation}

\section{Simulation Experiment}
\subsection{Simulation Setup}
\begin{figure}[!t]
\centering
\subfloat[]{\includegraphics[width=3.3in]{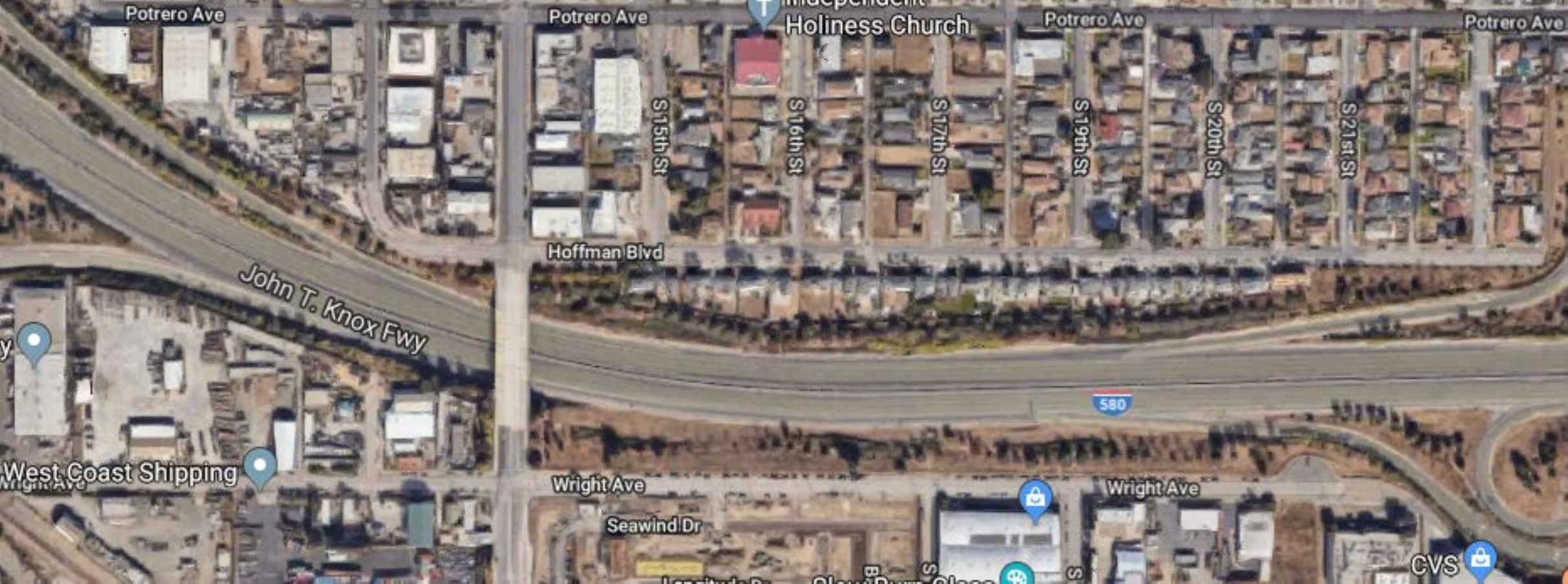}}\\
\subfloat[]{\includegraphics[width=3.3in]{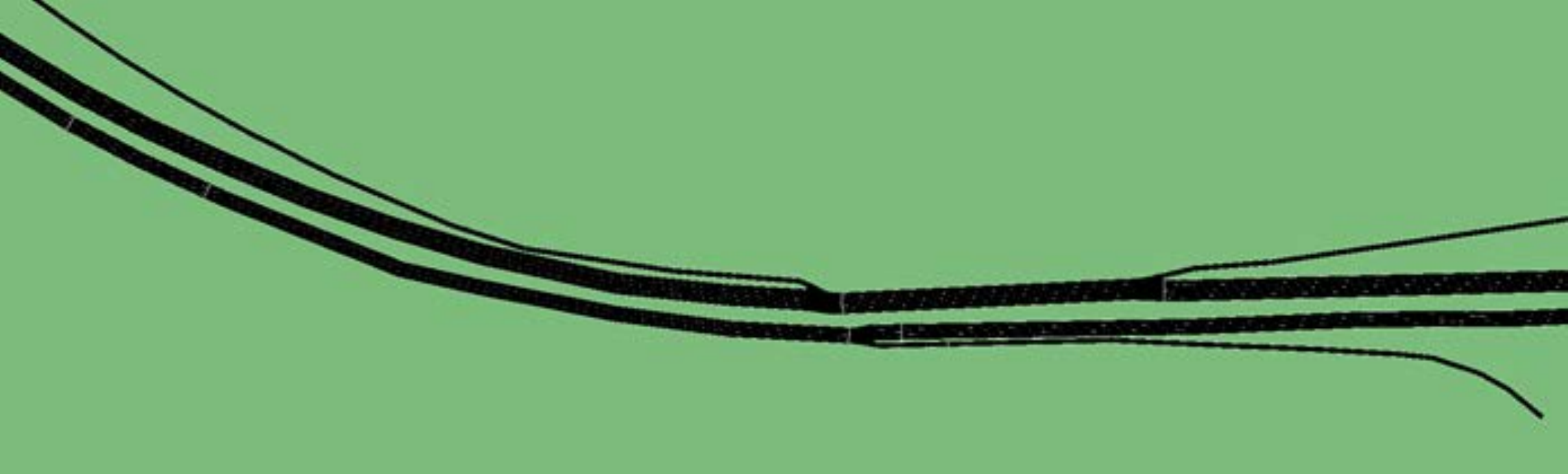}}
\caption{Simulation Network}
\label{fig:Google}
\end{figure}

The simulation network is modeled using a real-world highway segment with on-ramps and off-ramps as shown in Fig. \ref{fig:Google}, which is implemented on SUMO. The highway segment length to the ramp exit is 800 m and each lane width is 3.75 m. Vehicle counts are generated from a probability model to simulate the dense traffic. Considering the vehicle occupant comfort, we constrain the maximum acceleration and emergency braking deceleration of vehicles as $2.5~\mathrm{m/s^2}$ and $-4.5~\mathrm{m/s^2}$, respectively. The distance threshold for near-collision penalty is set to 10 m. 

In order to make the proposed simulation network as similar to real traffic as possible, we applied the intelligent driver model (IDM) to the other vehicles’ longitudinal control method as well as low-level longitudinal control of the ego-vehicle. In SUMO environment, we also designed the “abort” lane change maneuvers to reduce the risks of collision. The lane change kinematics in SUMO is explored and tested in both normal situations and challenging situations. When facing challenging situation such as in dense traffic, the vehicle can abort lane change anytime to avoid collisions and resume the lane changing maneuver when it’s safe to move. 

In terms of PPO hyper-parameters, we choose to use Adam and learning rate annealing with a step size of $1\times 10^{-4}$, and we set the horizon $T=512$, the mini-batch size as 64, the discount factor $\gamma=0.99$, and the clipped parameter $\epsilon=0.2$. More detailed simulation parameter setting is shown in Table \ref{tab:parameters}.
\begin{figure}[!t]
\centering
\subfloat[Longitudinal speed]{\includegraphics[width=2.8in]{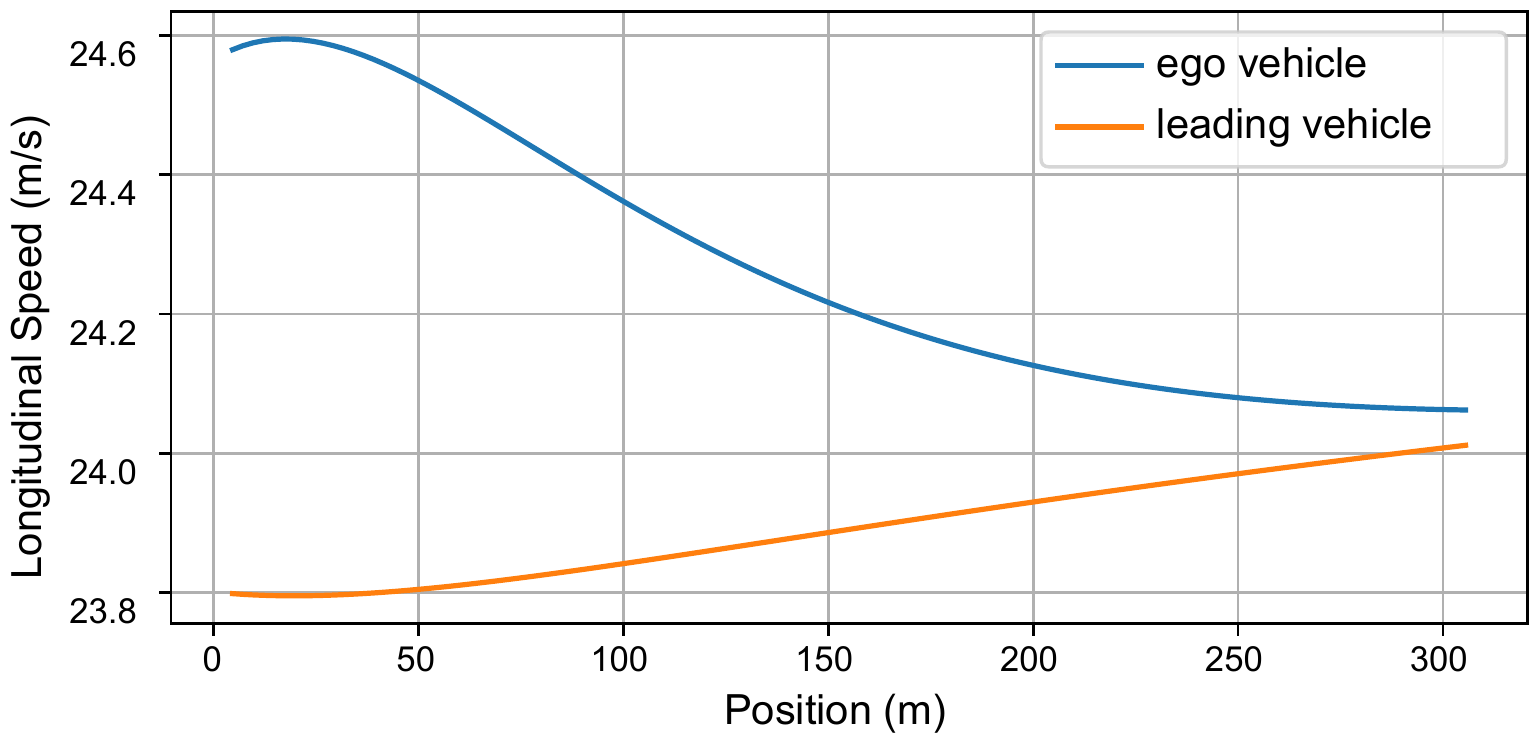}}\\
\subfloat[Distance to leader]{\includegraphics[width=2.8in]{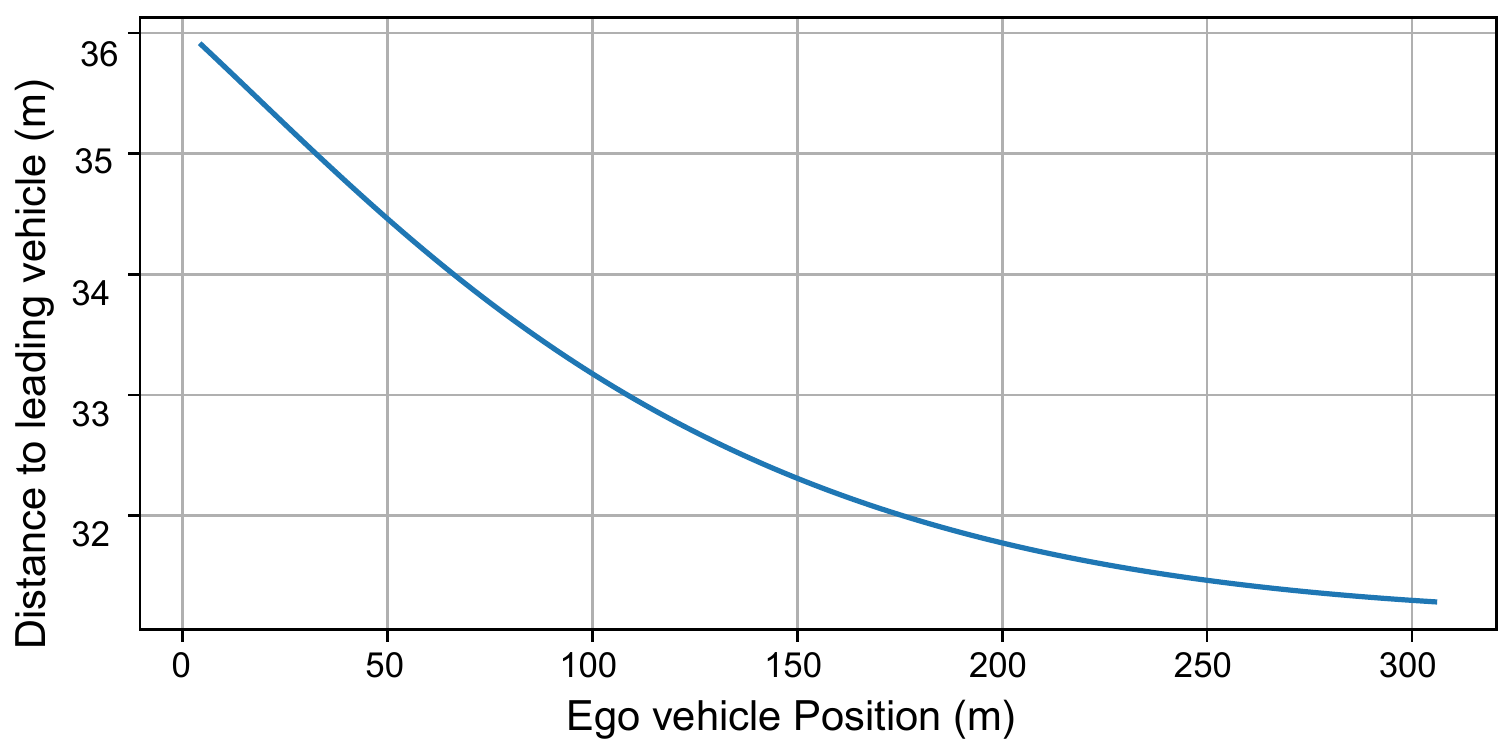}}
\caption{Longitudinal speed and distance to the leading vehicle.}
\label{fig:speed}
\end{figure}
%
%

%
%

\subsection{Evaluation Metrics}
To quantify the safety and effectiveness of the proposed PPO-based automated lane change model in both training and testing process, we add two metrics that evaluate the average collision rate and task success rate, which are computed under 10 rollouts with horizon $T$ being 512 in the training phase and 50 rollouts with horizon $T$ being 1024 in the testing phase, respectively. The collision counts both rear-end collisions and side-impact collisions. A successful task in a simulation run is defined as the ego-vehicle having successfully changed to the target lane before reaching the exit and managed to avoid collisions with other vehicles. 
\subsection{Results}
\begin{figure*}[!t]
\centering
\captionsetup[subfigure]{oneside,margin={0.4in,0in}}
\subfloat[]{\includegraphics[width=2.1in]{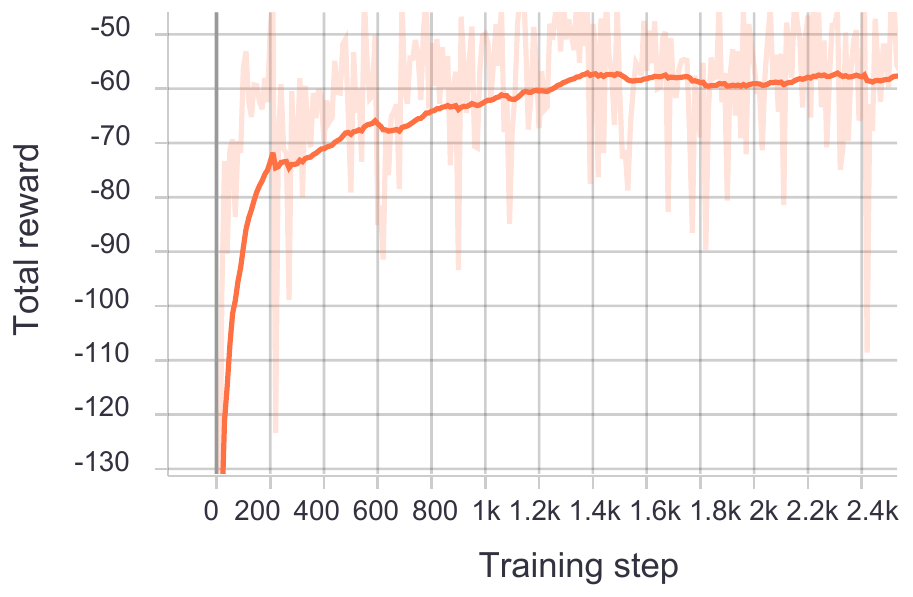}}
\hspace{0.2in}
\subfloat[]{\includegraphics[width=2.1in]{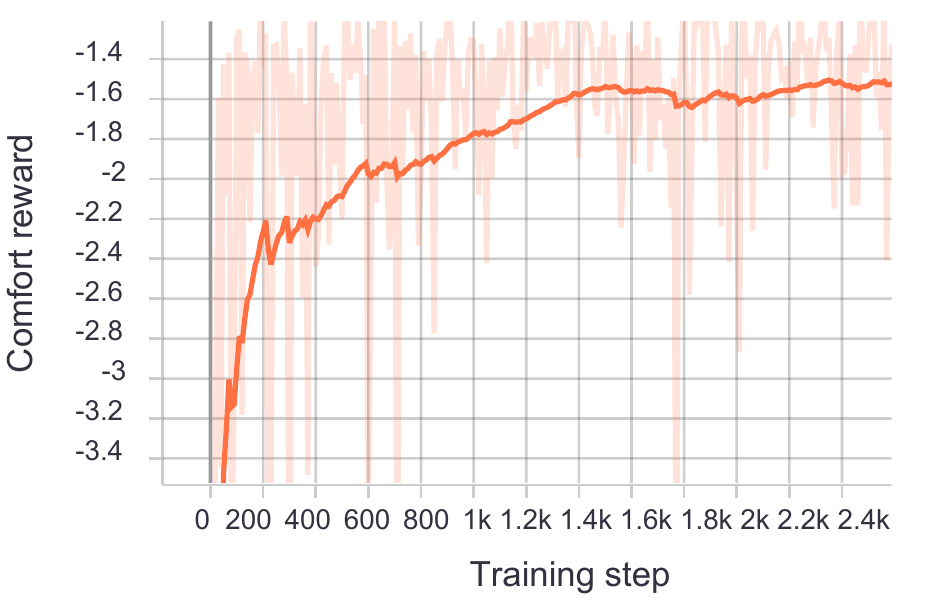}}
\\
\subfloat[]{\includegraphics[width=2.1in]{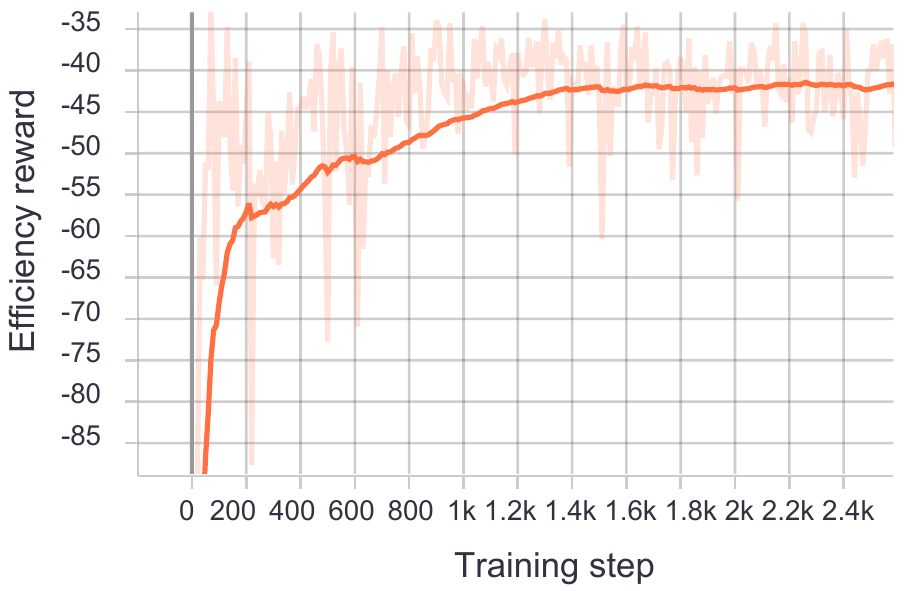}}
\hspace{0.2in}
\subfloat[]{\includegraphics[width=2.1in]{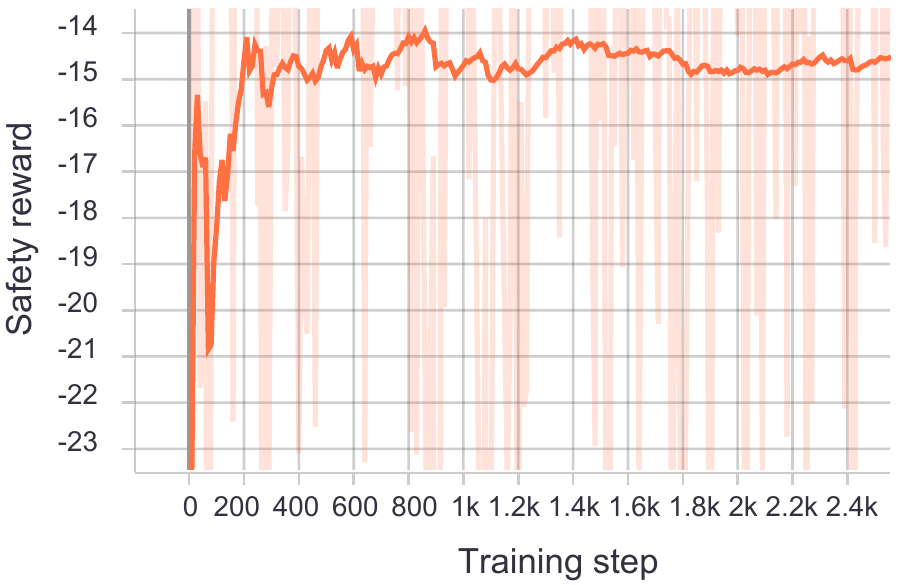}}
\caption{Learning curve of cumulative reward.}
\label{fig:reward}
\end{figure*}
\begin{figure*}[!t]
\centering
\captionsetup[subfigure]{oneside,margin={0.4in,0in}}
\subfloat[]{\includegraphics[width=2.1in]{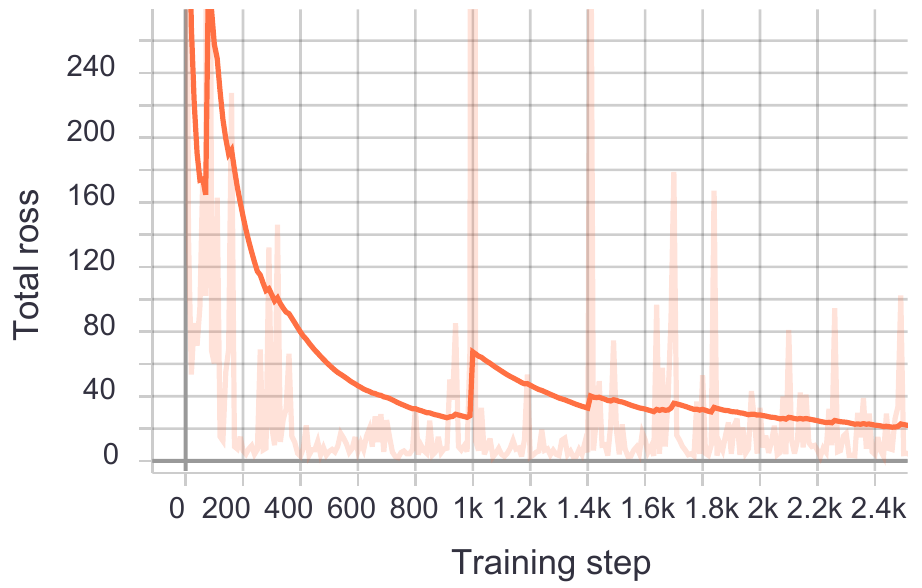}}
\hspace{0.2in}
\subfloat[]{\includegraphics[width=2.1in]{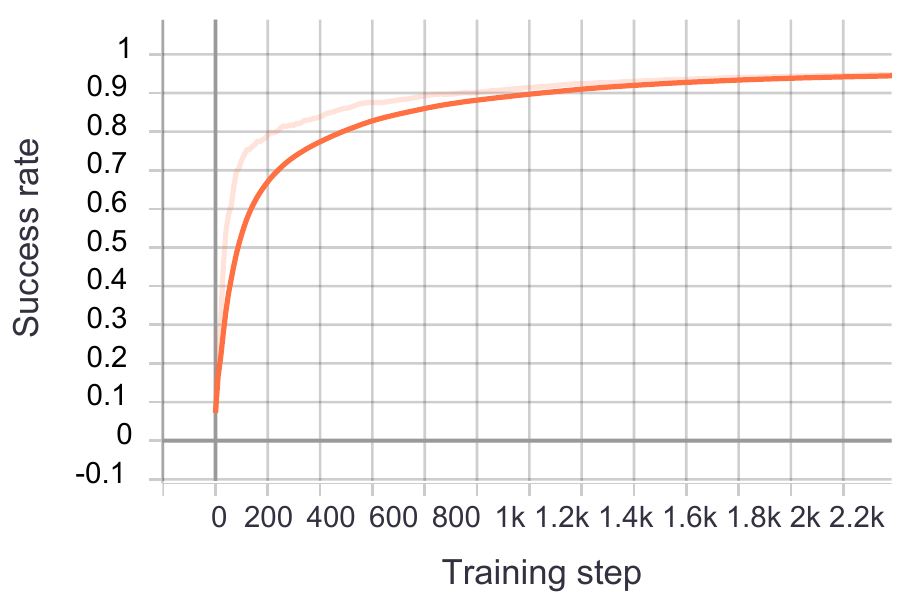}}
\hspace{0.2in}
\subfloat[]{\includegraphics[width=2.1in]{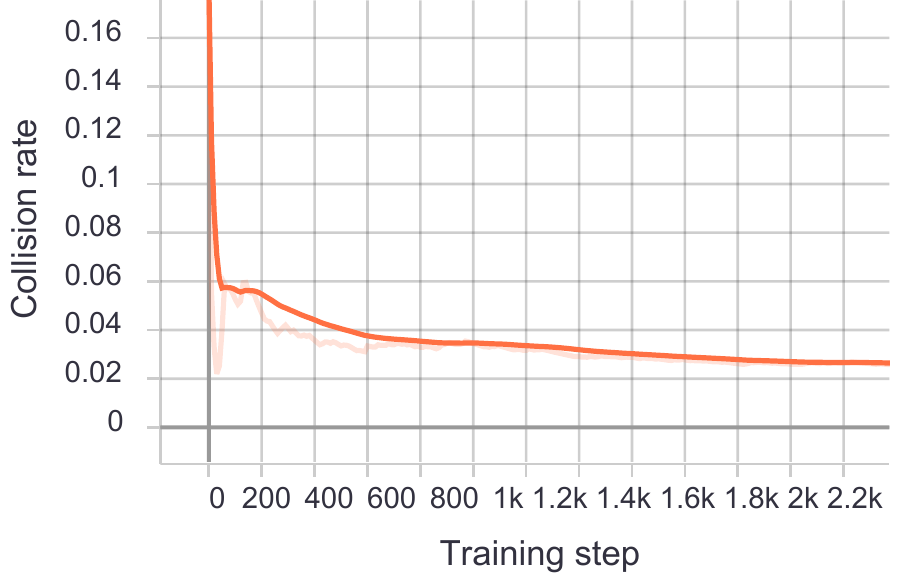}}
\caption{Total loss and evaluation results including success rate and collision rate.}
\label{fig:results}
\end{figure*}
%
%


In Fig. \ref{fig:speed}, we illustrate that the ego-vehicle can adjust its own speed to narrow the distance between itself and its leading vehicle after adopting the IDM model \cite{IDM}. If a decision is made to change lane and follow the target lane leader, the ego-vehicle then proceeds to accomplish this lane change decision. 

The average cumulative total reward, comfort reward, efficiency reward and safety reward in the training process are shown in Fig. \ref{fig:reward}. The curve of the cumulative reward indicates the ego-vehicle can successfully learn to take actions to maximize the reward for both cases. All three categories of reward curves (comfort, efficiency, safety) show an increasing trend. It is noted that a drop is observed in the safety reward at the early learning stage, illustrating the process balancing exploration and exploitation. 

Fig. \ref{fig:results} presents the total loss, cumulative collision rate and task success rate in the training process. In Fig. \ref{fig:results} (a), we can observe a clear decreasing trend in the average cumulative loss, which indicates the total loss converges after taking sufficient epochs. In the training process, a continuously increasing success rate can be observed in Fig. \ref{fig:results} (b). On the contrary, the collision rate in Fig. \ref{fig:results} (c) reaches its peak value at the early stage of training process, which is in accordance to the decrease of safety reward. Then it continuously decreases and shows an improving lane change performance in terms of safety. During the training process, the trained agent can achieve a 95\% success rate, while the collision rate is reduced from 20\% to 2\% in dense traffic. These curves indicate the ego-vehicle is capable of learning the mandatory lane change strategy with regard to our designed reward function using the proposed safe PPO model. 

In the testing scenarios, the trained agent with safety intervention that replacing the predicted catastrophic action can achieve 0.5\% average collision rate and 99.15\% average success rate over 50 rollouts with  horizon $T$ being 1024. For comparison, we implemented two versions of rule-based agent drive on the same test scenarios and action spaces that we created and measured their average return, collision rate and success rate as well. The trained rule-based agent utilize the safety gap between vehicles and time-to-collision (TTC) as criterion for decision making and we picked their best performance for comparison with tuning the threshold parameter. The average collision rate for rule-based agent with safety gap threshold set as 10 meters can reach 7.14\% with success rate 90.51\%. We observed with a further increase of the safety gap threshold, the collision rate can be reduced to zero but the success rate also dramatically decreases to below 70\%. The best performance of rule-based agent with TTC as criterion has presented with the average collision rate 6.31\% and with success rate 92.62\% respectively. The comparative testing results show the proposed safe PPO-based agent has promising high success rate while remaining the collision rate to close to zero.  

\begin{table}[]
\caption{Simulation parameters setup}
\centering
\resizebox{\linewidth}{!}{
\begin{tabular}{lll}
\toprule
Parameter & Value & Description                \\
\midrule
timesteps per actorbatch & 512 & number of steps of environment per update                \\
clip parameter           & 0.2  & clipping range                                           \\
optim epochs             & 10    & number of training epochs per update                     \\
learning rate            & 1e-4 & learning rate                                            \\
$\gamma$    & 0.99 & discounting factor                        \\
$\lambda$   & 0.95 & advantage estimation discounting factor   \\
\bottomrule
\end{tabular}}
\label{tab:parameters}
\end{table}
\begin{table}[]
\caption{Testing Results}
\centering
\begin{tabular}{lccc}
\toprule
              & average return      & collision rate (\%) & success rate (\%)      \\
\midrule
Safe\_PPO     & -38.54 & 0.50            & 99.15           \\
Baseline\_TTC & -45.40 & 6.31             & 92.62          \\
Baseline\_gap & -47.86 & 7.14            & 90.51          \\
\bottomrule
\end{tabular}
\label{tab:testing}
\end{table}

\section{Conclusions and Future Work}
This paper proposed an automated mandatory lane change strategy by using proximal policy optimization (PPO) \cite{PPO} based deep reinforcement learning, which features \emph{safety}, \emph{efficiency}, and \emph{comfort}. The high-level PPO policy is used to generate lane-change decisions (i.e. when and how) at each time step based on the current driving situations of the ego vehicle and its surrounding vehicles, while the lower level control is executed by a pre-defined model. We have shown the ego-vehicle trained using PPO based deep reinforcement learning can take appropriate actions to maximize the accumulated reward and achieve 95\% - 99\% success rate in dense traffic, which demonstrates the effectiveness of the proposed lane change strategy. 

The next step of our study is to incorporate human/expert inference with probability modeling to achieve better performance. Future research directions may also involve more interdisciplinary work of RL intertwined with other techniques for better generalization, some recent examples can be found in combining transfer learning \cite{DDQN-18, Zeroshot-18}.







\balance
\bibliographystyle{IEEEtran}
\bibliography{IEEEabrv.bib,ref.bib} 

\begin{thebibliography}{10}
\providecommand{\url}[1]{#1}
\csname url@samestyle\endcsname
\providecommand{\newblock}{\relax}
\providecommand{\bibinfo}[2]{#2}
\providecommand{\BIBentrySTDinterwordspacing}{\spaceskip=0pt\relax}
\providecommand{\BIBentryALTinterwordstretchfactor}{4}
\providecommand{\BIBentryALTinterwordspacing}{\spaceskip=\fontdimen2\font plus
\BIBentryALTinterwordstretchfactor\fontdimen3\font minus
  \fontdimen4\font\relax}
\providecommand{\BIBforeignlanguage}[2]{{%
\expandafter\ifx\csname l@#1\endcsname\relax
\typeout{** WARNING: IEEEtran.bst: No hyphenation pattern has been}%
\typeout{** loaded for the language `#1'. Using the pattern for}%
\typeout{** the default language instead.}%
\else
\language=\csname l@#1\endcsname
\fi
#2}}
\providecommand{\BIBdecl}{\relax}
\BIBdecl

\bibitem{NVIDIA}
M.~Bojarski, D.~Del~Testa, D.~Dworakowski, B.~Firner, B.~Flepp, P.~Goyal, L.~D.
  Jackel, M.~Monfort, U.~Muller, J.~Zhang \emph{et~al.}, ``End to end learning
  for self-driving cars,'' \emph{arXiv preprint arXiv:1604.07316}, 2016.

\bibitem{finn2016guided}
C.~Finn, S.~Levine, and P.~Abbeel, ``Guided cost learning: {D}eep inverse
  optimal control via policy optimization,'' in \emph{International Conference
  on Machine Learning}, 2016, pp. 49--58.

\bibitem{Atari2013}
V.~Mnih, K.~Kavukcuoglu, D.~Silver, A.~Graves, I.~Antonoglou, D.~Wierstra, and
  M.~Riedmiller, ``Playing atari with deep reinforcement learning,'' 2013.

\bibitem{wang2019continuous}
P.~Wang, H.~Li, and C.-Y. Chan, ``Continuous control for automated lane change
  behavior based on deep deterministic policy gradient algorithm,'' in
  \emph{Proc. IEEE Intell. Veh. Sympo. (IV)}, June 2019, pp. 1454--1460.

\bibitem{review-17}
A.~Sallab, M.~Abdou, E.~Perot, and S.~Yogamani, ``Deep reinforcement learning
  framework for autonomous driving,'' \emph{Electronic Imaging}, vol. 2017,
  no.~19, p. 70–76, Jan 2017.

\bibitem{wang2019quadratic}
P.~Wang, H.~Li, and C.-Y. Chan, ``Quadratic {Q}-network for learning continuous
  control for autonomous vehicles,'' \emph{NIPS Workshop Mach. Learn. Auton.
  Driving}, 2019.

\bibitem{Collision1}
G.~{Xu}, L.~{Liu}, Y.~{Ou}, and Z.~{Song}, ``Dynamic modeling of driver control
  strategy of lane-change behavior and trajectory planning for collision
  prediction,'' \emph{IEEE Transactions on Intelligent Transportation Systems},
  vol.~13, no.~3, pp. 1138--1155, Sep. 2012.

\bibitem{Collision2}
D.~Yang, S.~Zheng, C.~Wen, P.~J. Jin, and B.~Ran, ``A dynamic lane-changing
  trajectory planning model for automated vehicles,'' \emph{Transp. Research
  Part C: Emerging Technol.}, vol.~95, pp. 228--247, 2018.

\bibitem{LHP}
F.~{Ye}, G.~{Wu}, K.~{Boriboonsomsin}, M.~J. {Barth}, S.~{Rajab}, and S.~{Bai},
  ``Development and evaluation of lane hazard prediction application for
  connected and automated vehicles (cavs),'' in \emph{2018 21st International
  Conference on Intelligent Transportation Systems (ITSC)}, Nov 2018, pp.
  2872--2877.

\bibitem{CooperativeLC-19}
G.~Wang, J.~Hu, Z.~Li, and L.~Li, ``Cooperative lane changing via deep
  reinforcement learning,'' \emph{arXiv preprint arXiv:1906.08662}, 2019.

\bibitem{QDRL-17}
M.~Mukadam, A.~Cosgun, A.~Nakhaei, and K.~Fujimura, ``Tactical decision making
  for lane changing with deep reinforcement learning,'' \emph{NIPS Workshop
  Mach. Learn. Int. Transp. Syst.}, 2017.

\bibitem{Pin-2018}
P.~Wang, C.-Y. Chan, and A.~d.~L. Fortelle, ``A reinforcement learning based
  approach for automated lane change maneuvers,'' \emph{IEEE Intell. Veh. Symp.
  (IV)}, Jun 2018.

\bibitem{Tianyu-19}
T.~{Shi}, P.~{Wang}, X.~{Cheng}, C.~{Chan}, and D.~{Huang}, ``Driving decision
  and control for automated lane change behavior based on deep reinforcement
  learning,'' in \emph{Proc. Int. Conf. Intell. Transp. Syst. (ITSC)}, Oct
  2019, pp. 2895--2900.

\bibitem{HighLC-18}
B.~{Mirchevska}, C.~{Pek}, M.~{Werling}, M.~{Althoff}, and J.~{Boedecker},
  ``High-level decision making for safe and reasonable autonomous lane changing
  using reinforcement learning,'' in \emph{Proc. Int. Conf. Intell. Transp.
  Syst. (ITSC)}, Nov 2018, pp. 2156--2162.

\bibitem{AutoDM-19}
X.~{Xu}, L.~{Zuo}, X.~{Li}, L.~{Qian}, J.~{Ren}, and Z.~{Sun}, ``A
  reinforcement learning approach to autonomous decision making of intelligent
  vehicles on highways,'' \emph{IEEE Trans. Syst., Man, Cybern. Syst}, pp.
  1--14, 2018.

\bibitem{Shield2017}
M.~Alshiekh, R.~Bloem, R.~Ehlers, B.~Könighofer, S.~Niekum, and U.~Topcu,
  ``Safe reinforcement learning via shielding,'' 2017.

\bibitem{PPO}
J.~Schulman, F.~Wolski, P.~Dhariwal, A.~Radford, and O.~Klimov, ``Proximal
  policy optimization algorithms,'' \emph{arXiv preprint arXiv:1707.06347},
  2017.

\bibitem{TRPO}
J.~Schulman, S.~Levine, P.~Abbeel, M.~Jordan, and P.~Moritz, ``Trust region
  policy optimization,'' in \emph{Proc. Int. Conf. Mach. Learn. (ICML)}, 2015,
  pp. 1889--1897.

\bibitem{SafeRL2017}
W.~Saunders, G.~Sastry, A.~Stuhlmueller, and O.~Evans, ``Trial without error:
  Towards safe reinforcement learning via human intervention,'' 2017.

\bibitem{Reinforce}
R.~J. Williams, ``Simple statistical gradient-following algorithms for
  connectionist reinforcement learning,'' \emph{Machine learning}, vol.~8, no.
  3-4, pp. 229--256, 1992.

\bibitem{CPI}
S.~Kakade and J.~Langford, ``Approximately optimal approximate reinforcement
  learning,'' in \emph{ICML}, vol.~2, 2002, pp. 267--274.

\bibitem{SUMO}
J.~E. M.~Behrisch, L.~Bieker and D.~Krajzewicz, ``Sumo--simulation of urban
  mobility: {A}n overview,'' pp. 63--68, 2011.

\bibitem{IDM}
M.~Treiber, A.~Hennecke, and D.~Helbing, ``Congested traffic states in
  empirical observations and microscopic simulations,'' \emph{Physical Review
  E}, vol.~62, no.~2, p. 1805–1824, Aug 2000.

\bibitem{DDQN-18}
C.-J. Hoel, K.~Wolff, and L.~Laine, ``Automated speed and lane change decision
  making using deep reinforcement learning,'' \emph{Proc. Int. Conf. Intell.
  Transp. Syst. (ITSC)}, Nov 2018.

\bibitem{Zeroshot-18}
Z.~Xu, C.~Tang, and M.~Tomizuka, ``Zero-shot deep reinforcement learning
  driving policy transfer for autonomous vehicles based on robust control,''
  \emph{Proc. Int. Conf. Intell. Transp. Syst. (ITSC)}, Nov 2018.

\end{thebibliography}
\balance

\end{document}